\documentclass[letterpaper, 10 pt, conference]{ieeeconf}

\IEEEoverridecommandlockouts                              

\usepackage{graphicx} 
\usepackage{amsmath} 
\usepackage{balance}
\usepackage{algorithm,algorithmic}

\title{\LARGE \bf
Path Optimization for Ground Vehicles in Off-Road Terrain
}

\author{Timothy Overbye and Srikanth Saripalli
\thanks{$^{1}$Timothy Overbye and Srikanth Saripalli are with the Department of Mechanical Engineering, Texas A\&M University, College Station, TX 77840, USA
        {\tt\small overbye2@tamu.edu}
        {\tt\small ssaripalli@tamu.edu}}%
}

\begin{document}

\maketitle
\thispagestyle{empty}
\pagestyle{empty}

\begin{abstract}

We present a method for path optimization for ground vehicles in off-road environments at high speeds. This path optimization considers the kinematic constraints of the vehicle. By thinking in the actuator space we can represent these constraints as limits in the space rather than derived properties of the path. In this paper we present a actuator space approach to path optimization for off-road ground vehicles. This is done by representing and operation on the path as a list of steering angles over the path length. This transforms the set of kinematic constraints into constraints on the steering angle. We then put this path into a gradient descent solver. This produced paths that are kinematically feasible and optimized in accordance with our cost function. Finally, we tested the system both in simulation and on an off-road vehicle at speeds of 5~m/s.
\end{abstract}

\section{INTRODUCTION}

High speed navigation requires paths that can be generated quickly, satisfy the constraints of the vehicle, and avoid collision. Unfortunately, the quality of the resulting path and the time to compute the path are often inversely related. To solve this issue hierarchical planners are often used that have shorter, but higher fidelity, trajectories at the lowest level. However, if the higher level paths contain infeasable paths then these lower level trajectories can easily get stuck. Therefore, we want a method to take these higher level paths and improve their feasibility before they're passed to the lower levels. Since a large challenge to feasibility is kinematic constraints it makes sense to transform the path to a representation that makes these constraints easy to satisfy. Here we use the actuator space representation. That is, rather than represent the path as a list of vehicle states we instead represent the path as a list of control inputs. The trade-off is an increase in complexity of world space constraints such as obstacles.

In this paper we propose a path optimization framework in this actuator space for ground vehicles. By optimizing the higher level path additional constraints can be applied to it after generation. Next we will discuss the benefits of both optimization and the actuator space representation in more detail.

   \begin{figure}[htpb]
      \centering
      \framebox{\parbox{3in}{
      
      \includegraphics[width=\linewidth]{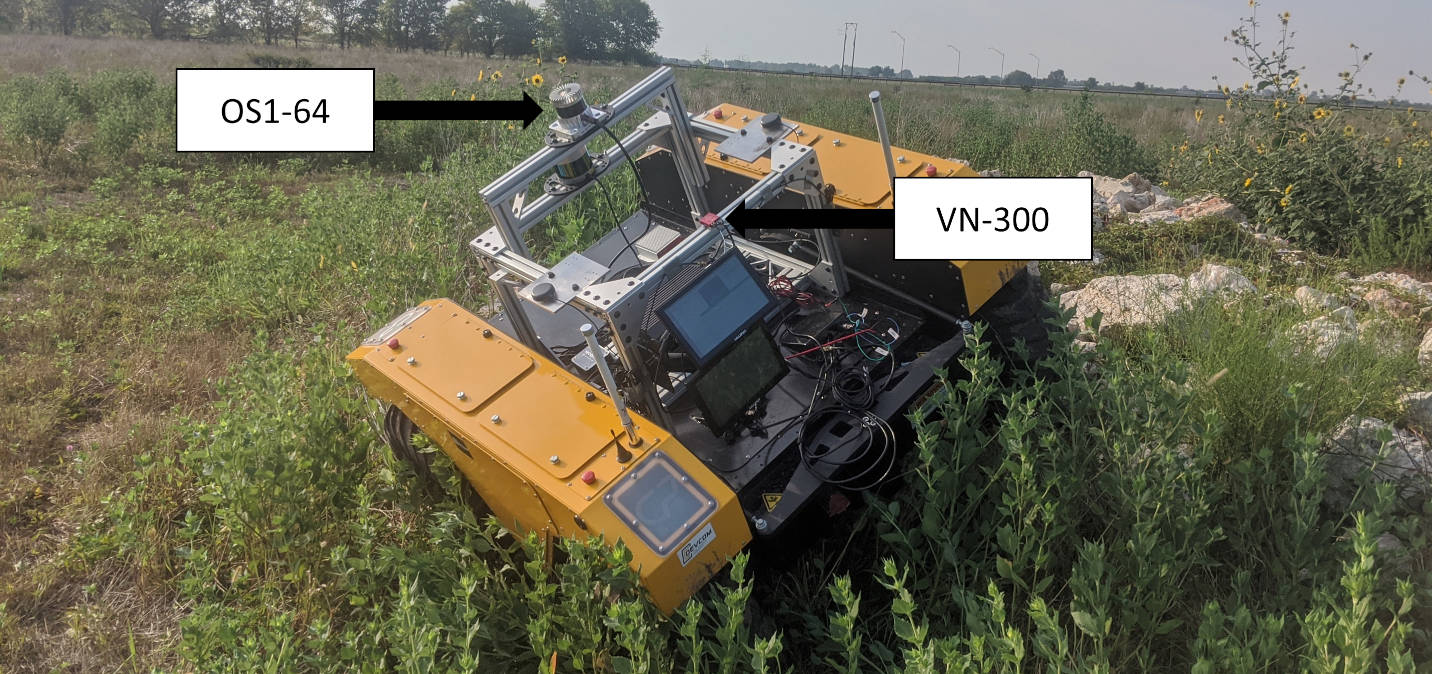}
	}}
      
      \caption{The Warthog with attached sensors. }
      \label{fig:warthog}
   \end{figure}
   
\subsection{Path Optimization}
Many methods to get optimal paths that satisfy all constraints are slow. This is important due to the relationship between processing rate, vehicle speed, and the minimum reaction radius. The minimum reaction radius is the minimum distance at which the vehicle is capable of reacting to new information. It can be thought of as the total system latency multiplied by the vehicle speed. Since path planning (along with sensor segmentation)  takes up the bulk of this time, this radius can be significantly reduced by improved planning methods. The ideal situation is a planner that is both fast and satisfies all constraints. However, this is hard to achieve~\cite{planningStudy}\cite{planningEvaluation}.
Another solution is to take a quickly generated path and apply constraints after generation through an optimization process. By doing this you can gain some of the benefits of both fast and accurate planners. Although it should be noted that the constraints may significantly change the optimal path such that a path optimizer, if given this initial path, will converge to some locally optimal path rather than the global optimal. Thus, a distinction should be made between an optimized path and an optimal path.

\subsection{Actuator Space Representation}

As vehicle speed increases the kinematic, and even dynamic, constraints can restrict the free space of the vehicle even more than the physical obstacles present in the environment. Therefore, it becomes advantageous to start thinking in a framework where these constraints can be nativity represented. Typical world space representation are good at representing obstacles but have no native way to represent kinematic constraints such minimum turning radius. Additionally, path properties like smoothness are harder to quantify. 

Representing the path in the actuator space switches this problem around. Now, provided with a good vehicle model, kinematic and smoothness constraints can be represented as obstacles in the actuator space. However, as vehicles inevitably operate in the real world, and typically aren't interested in purely achieving some actuator state, we must still consider the world space. Thus, requiring us to transform between representations when checking for obstacles. This leads to a trade-off between the ease of representing kinematic constraints and world constraints.

\section{RELATED WORK}

There have been many works published on the path planning and optimization problem. Here, we will review some of them as they pertain to some related domains.

\subsection{Manipulators}

This concept of actuator space planning has a long history in manipulator planning. In fact, it's typical for manipulator planning to be done entirely within the actuator space with obstacles transformed into it from the world space. However, manipulators have the advantage that their actuator space and state space are identical. Typically, they are trying to achieve given state in the actuator space so, aside from the initial obstacle transform, transforming states isn't necessary. This leads to an ideal environment for optimization with several works being done~\cite{chomp}\cite{ArmOptimization1}\cite{graspingAsOptimization}.

\subsection{Air Vehicles}

Recently, the concept of actuator space has had some applications in planning for air vehicles, typically small quadcoptor Unmanned Air Systems (UAS). However, this transform will often only be used to check that the path is within the limits of the actuator\cite{Richter2016quadbezier}\cite{dronesmoothing}. Due to the kinematics of these UAS, scaling the path speed around offending segments is typically all that's needed to satisfy these constraints. So the path can be transformed into the actuator space and rescaled without the need to directly plan in the space.

From an optimization point of view, UAS are another good candidate for path optimization~\cite{droneContinuousOptimization}\cite{DONG2019DroneSmoothing}. They typically operate in binary space, that is, the space is either free or has an obstacle and there is no preference for one set of free space over another. Additionally, due to their higher speed, path smoothness is highly desired. Thus making kinematic constraints the dominating factor in path planning.

\subsection{Ground Vehicles}
For ground vehicles, thinking in the actuator space isn't a new idea. Model predictive control can be thought of as operating in this context. Additionally, many methods of trajectory sampling or path primitive generation fall into this context. However, due to the larger set of constraints on ground vehicles and the comparatively complex set of world space obstacles, longer path planning is typically done in the world space. Although there have been some examples of actuator space planning for short trajectories~\cite{shortRangeOptimization}.

Optimization has also seen comparatively less literature than in the other domains. Most of what exists is used as a path smoother for planners such as the RRT family~\cite{SmoothingReview}. The history of path optimizers can be thought of as beginning with the application of Dubins~\cite{dubins} (and later Reed-Sheep's paths) to methods such as A*~\cite{aStar}\cite{lattice1} and it's derivatives, such as D*~\cite{DStar} and D*~Lite~\cite{DStarLite}, and RRT~\cite{RRT}. Although these paths can be thought of as satisfying all the kinematic constraints they are discontinuous in steering angle. 
A solution to this is the clothoid path. Clothoid guarantee continuous steering angle and limits such as max steering angle and max steering angle rate are natural to the path formulation\cite{vanderMolen1992clothoids}. However, there is no closed form solution to the clothoid path. Additionally, clothoids can only represent a subset of all feasible paths. 
Most recent optimization work has represented the paths as polynomial splines \cite{ConstrainedOptimization}\cite{ELHOSENY2018Bezier}\cite{MAEKAWA2010splines}\cite{bSplineSmoothing}. These splines trade some of the benefits of clothoids for the closed form nature of polynomials. Additionally, they can represent a much larger segment of the set of all feasible paths.

\section{APPROACH}
\begin{figure}[htpb]
      \centering
      \framebox{\parbox{3in}{
      
      \includegraphics[width=\linewidth]{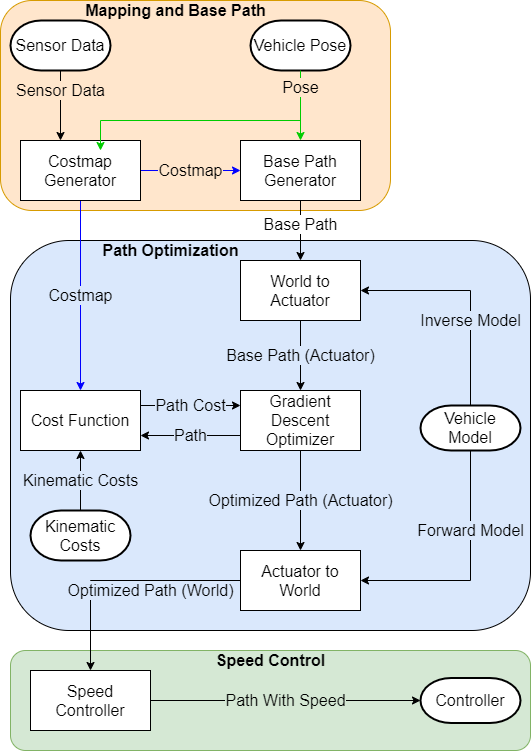}
	}}
      
      \caption{Optimizer architecture. From the top, sensor and position data is used to make the costmap and base path. In the middle is the path optimizer. Finally, the speed controller at the bottom}
      \label{fig:opt_architecture}
   \end{figure}

Our approach to this problem can be divided into three main stages as shown in figure~\ref{fig:opt_architecture}. First is costmap and base path generation. Here sensor data is fused into a costmap and an initial path is made over the map. Next in the optimizer this path is fist transformed into the actuator space. The path is put into an iterative gradient descent solver until either a minimum cost is achieved or a maximum number of iterations is reached. Finally, we reach the speed control section where the resulting path is put through a speed controller assigning each point along the path a target speed.

\subsection{Cost Map Creation}

The cost map process is a continuation on the methods developed in~\cite{myPaper}. A lidar scan is taken over a local grid map. From this scan, obstacles are extracted by taking the heigh difference between the minimum and maximum height return in each cell. If this difference is larger than a given threshold then the cell is marked as containing an obstacle. Here, we consider two types of obstacles, hard obstacles (trees, rocks, people, etc) are those that should never be driven through, and soft obstacles (bushes, tall grass, etc) that should be avoided but are traversable. This is determined by our segmentation layer, each lidar return is marked as either a hard or soft object. If there are soft object returns above the obstacle threshold but no hard returns then the cell is a soft obstacle. Otherwise, if there's a hard return above the threshold it's a hard obstacle.

\begin{figure}[htpb]
      \centering
      \framebox{\parbox{3in}{
	  \includegraphics[width=\linewidth]{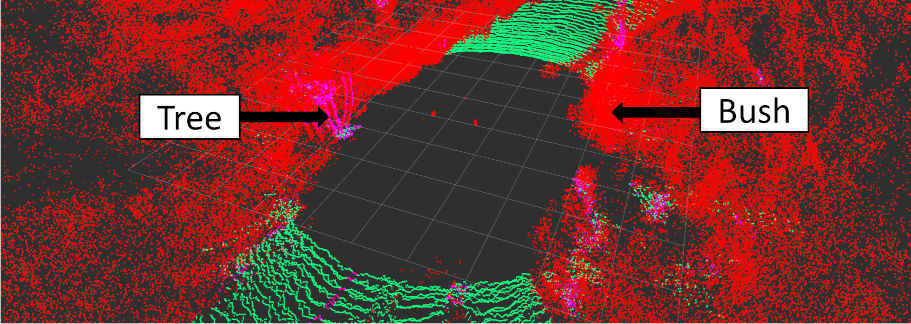}
      \includegraphics[width=\linewidth]{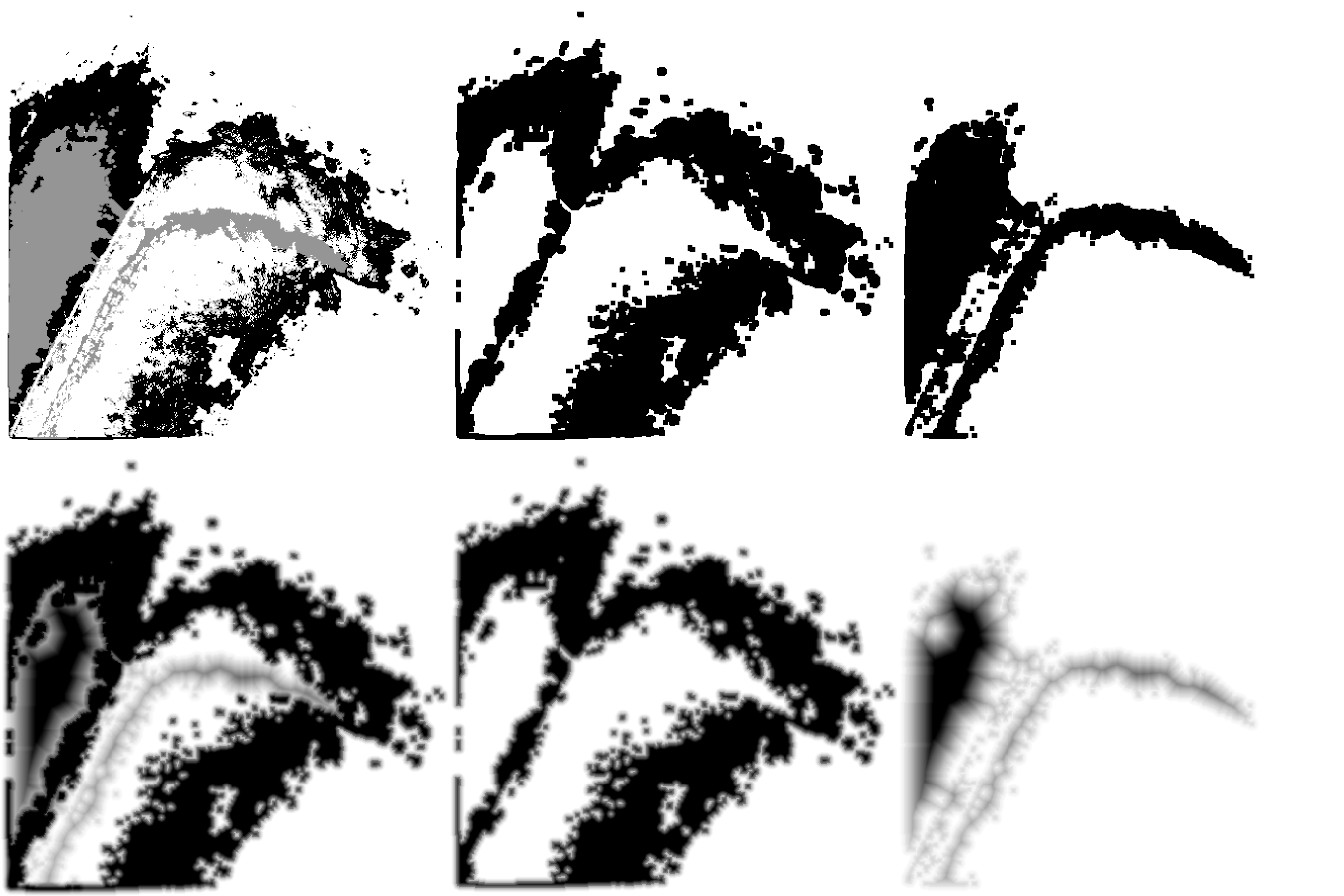}
      
	}}
      
      \caption{Cost map formulation. Top, a sample segmented lidar scan with hard obstacles in pink,  soft obstacles in red, and ground in green. Middle left, the input map with hard obstacles in black, soft obstacles in grey, and free space in white. The expanded hard and soft obstacle maps (center middle ,center right). The cost gradients associated with both obstacle maps (bottom middle ,bottom right). Finally, the combined cost map (bottom left).}
      \label{fig:cost_map_grad}
   \end{figure}

Both the hard and soft obstacle maps share a common process, shown in figure~\ref{fig:cost_map_grad}. As they are binary maps, rather than a continuous function, they can be expanded with a dilate operation by the radius of the robot. The area inside this expanded region is then converted into a distance region. Where the value of each pixel represents the distance to the edge of the region. This is done to ensure that there is always a strong gradient in the costmap to push the solution out of obstacles. Finally, a similar process is done to create a repulsive halo around around obstacles. This, once again, helps the solver avoid obstacles but also prevents discontinuities in the cost map. 

The final costmap is the simple per-pixel sum of both of these sub-maps. Thus, we maintains both the strong gradients and continuity in the final costmap.

\subsection{Base Path Generation}

Before the optimizer can run it needs to be given a base path. For our implementation we chose to use A*~\cite{aStar}. It was chosen due to its guarantee of optimality and fast run time. Here, we used 8 way connectivity and the euclidean distance as a heuristic. The vehicle's current position is used as the start of the search. If the goal point is within the local map it is used as the goal for A*. Otherwise, the closest point within the map to the goal is used for A*. This search is done over the same costmap as the optimization. Despite the guarantee of optimality of A* it doesn't take into account the kinematic constraints of the vehicle. Thus the final optimized path will deviate from the A* path.

\subsection{Path Representation}

\begin{figure}[htpb]
      \centering
      \framebox{\parbox{3in}{
      
      \includegraphics[width=\linewidth]{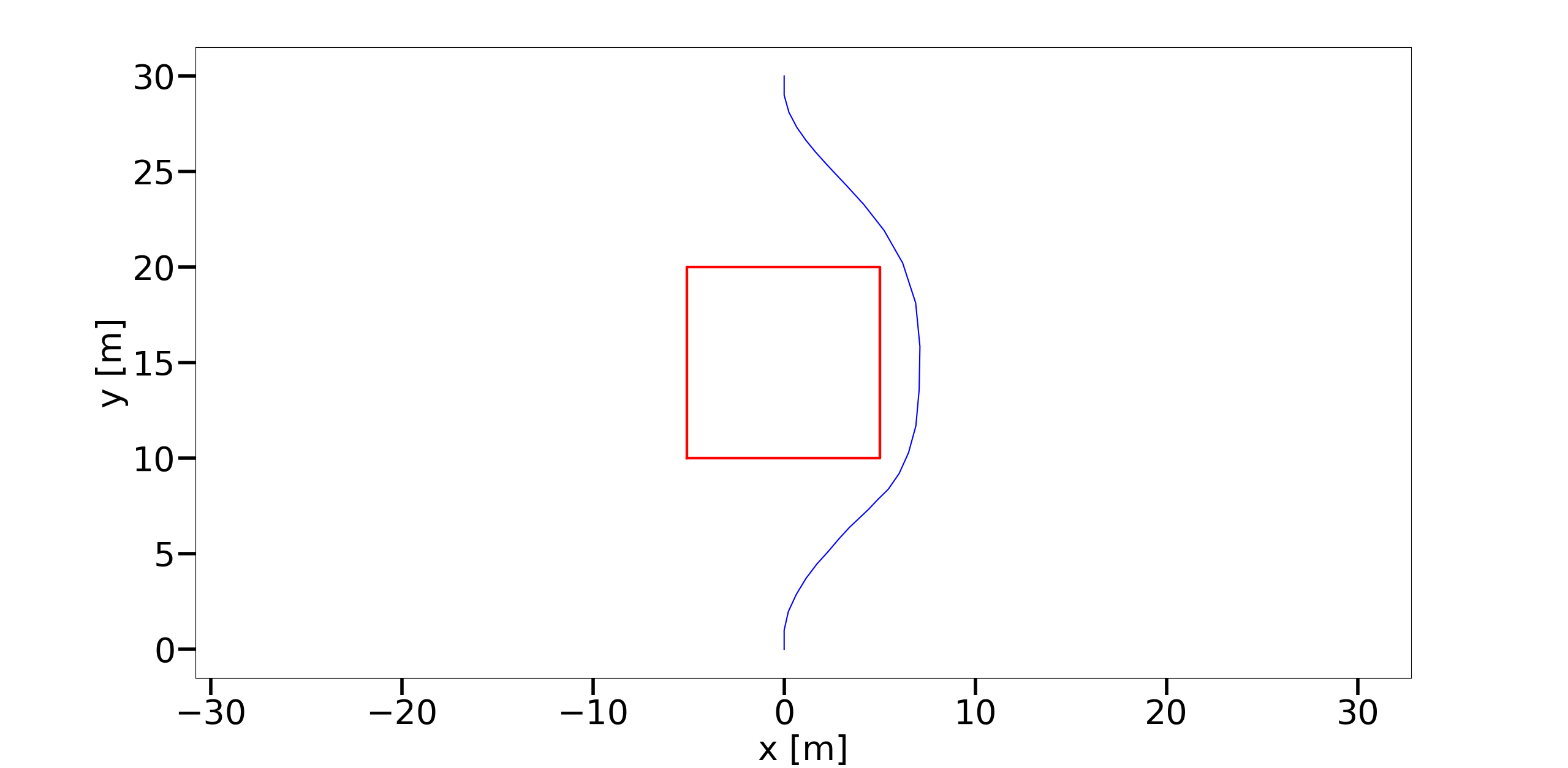}
      \includegraphics[width=\linewidth]{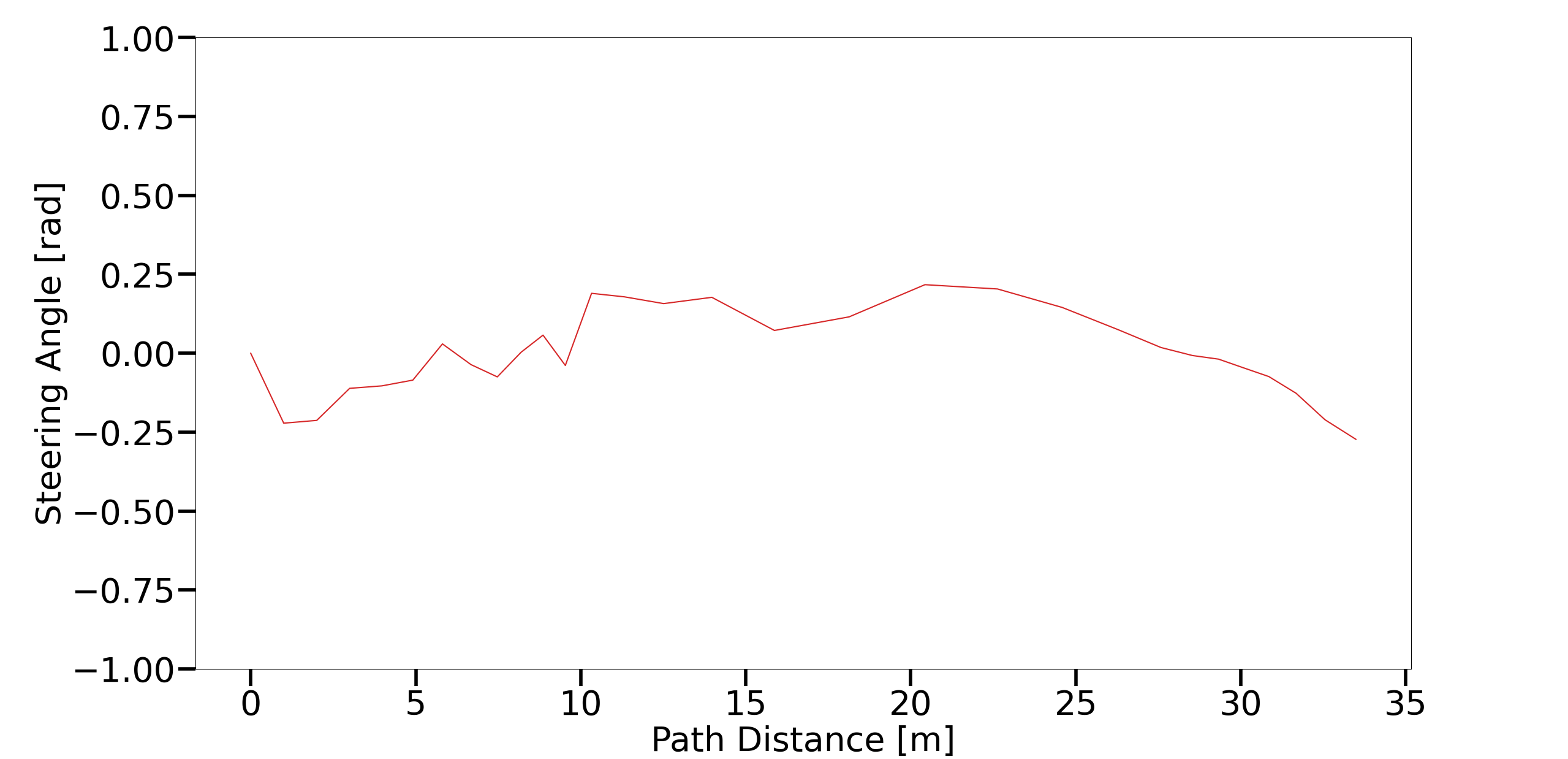}
	}}
      
      \caption{A path around a square obstacle represented in both world space (top) and actuator space (bottom).}
      \label{fig:path_representation}
   \end{figure}

Once the base path is generated it needs to be converted into the actuator space. The typical path representation in the world space is a list of $(x, y, \theta)$ representing the position and orientation of the vehicle on a 2D plane. An actuator space path is represented by a list of $(\phi, dr)$ along with the initial $(x_0, y_0, \theta_0)$. Here $\phi$ is the current steering angle and $dr$ is the distance traveled over some fixed time step. We assume a constant path speed. Figure~\ref{fig:path_representation} shows an example path represented in both world and actuator space.

To convert the path from world space to actuator space we need to have an inverse model of the vehicle. Note that this method isn't tied to a particular model but does require an inverse model. Here we use a 2D model although a 3D model could be used at the cost of increased dimensionality. Equations \ref{equ:inverse_phi} and \ref{equ:inverse_dr} describe the inverse model. If the input path doesn't contain heading information then we can use equation~\ref{equ:heading_from_path} to back calculate it. Although this will leave the heading at the final point undefined, in practice it can be assumed to be the same as that of the next to last point.

\begin{equation}
\label{equ:inverse_phi}
\phi_{i} = \theta_{i+1} - \theta{i}
\end{equation}
\begin{equation}
\label{equ:inverse_dr}
dr = \sqrt{(x_{i+1} - x_{i})^2 + (y_{i+1} - y_{i})^2 }
\end{equation}
\begin{equation}
\label{equ:heading_from_path}
\theta_{i} = \arctan(y_{i+1} - y_{i},x_{i+1} - x_{i} )
\end{equation}

Transforming from the actuator space to world space is easier. It can be thought of as simply integrating the forward model with respect to some starting point $(x_0,y_0,\theta_0)$. Equations~\ref{equ:forward_model_th},~\ref{equ:forward_model_x}, and~\ref{equ:forward_model_y} describe this process.

\begin{equation}
\label{equ:forward_model_th}
\theta_{i+1} = \theta_{i} + \phi_{i}
\end{equation}
\begin{equation}
\label{equ:forward_model_x}
x_{i+1} = x_{i} + \cos(\theta_{i})dr
\end{equation}
\begin{equation}
\label{equ:forward_model_y}
y_{i+1} = y_{i} + \sin(\theta_{i})dr
\end{equation}

\subsection{Cost Function Formulation}

Finally, we define the cost function to optimize against (equation~\ref{equ:cost_function1}). Here $P$ represents the path and $p$ is a point in the path. We have already talked about the costmap but to fully capture the cost of a path we must also consider it's kinematic properties. This leads to a cost function with both a map cost and a kinematic cost.

\begin{equation}
\label{equ:cost_function1}
F(P) =  F_{map}(P) + F_{path}(P)
\end{equation}

Breaking down this function further leads to the next two equations. The costmap cost (equation~\ref{equ:cost_function_map}) is simply the sum of all the cost map cells through which the path moves.

\begin{equation}
\label{equ:cost_function_map}
F_{map}(P) = \sum_{p}^{P} Map[p_x,p_y]
\end{equation}

Next, we are concerned with the constraints imposed on the vehicle by our kinematic model. Here we get two main constraints on the path, the steering angle $\phi$ must lie between some maximum and minimum values (equation~\ref{equ:constraint_angle_1}). Typically, we can assume that $-\phi_{min} = \phi_{max}$ so this constraint reduces to equation~\ref{equ:constraint_angle_2}.

\begin{equation}
\label{equ:constraint_angle_1}
\phi_{min} \leq \phi \leq \phi_{max}
\end{equation}
\begin{equation}
\label{equ:constraint_angle_2}
\vert \phi \vert \leq \phi_{max}
\end{equation}

Similarly, the rate of change in the steering angle is also of interest to us. This leads to another set of constraints in equation~\ref{equ:constraint_angle_rate_1} . Once again we can assume that $-\dot{\phi}_{min} = \dot{\phi}_{max}$ so this constraint reduces to equation~\ref{equ:constraint_angle_rate_2}.

\begin{equation}
\label{equ:constraint_angle_rate_1}
\dot{\phi}_{min} \leq \dot{\phi} \leq \dot{\phi}_{max}
\end{equation}
\begin{equation}
\label{equ:constraint_angle_rate_2}
\vert \dot{\phi} \vert \leq \dot{\phi}_{max}
\end{equation}

Rather than writing these as hard constraints, we instead represent them by large, steep cost gradients outside of the boundaries. However, even within the boundaries imposed by our constraints we are still interested in smooth paths. Equation~\ref{equ:cost_function_path} is the final kinematic cost function with $\alpha, \beta$ being the in and out of bounds cost factors respectively.

\begin{equation} 
\label{equ:cost_function_path}
F_{path}(P) = \sum_{p}^{P}( f( p_{\phi}, \alpha_{\phi},\beta_{\phi} ) + f( p_{\dot{\phi}}, \alpha_{\dot{\phi}},\beta_{\dot{\phi}} ) )
\end{equation}
\[
    f(\phi,\alpha,\beta) =
\begin{cases}
    \alpha \vert \phi \vert, & \text{if } \vert \phi \vert < \phi_{max}\\
    \alpha \phi_{\max} + \beta ( \vert \phi \vert - \phi_{\max}) ,             & \text{otherwise}
\end{cases}
\]

\subsection{Path Solver}

\begin{figure}[htpb]
      \centering
      \framebox{\parbox{3in}{
      	\includegraphics[width=\linewidth]{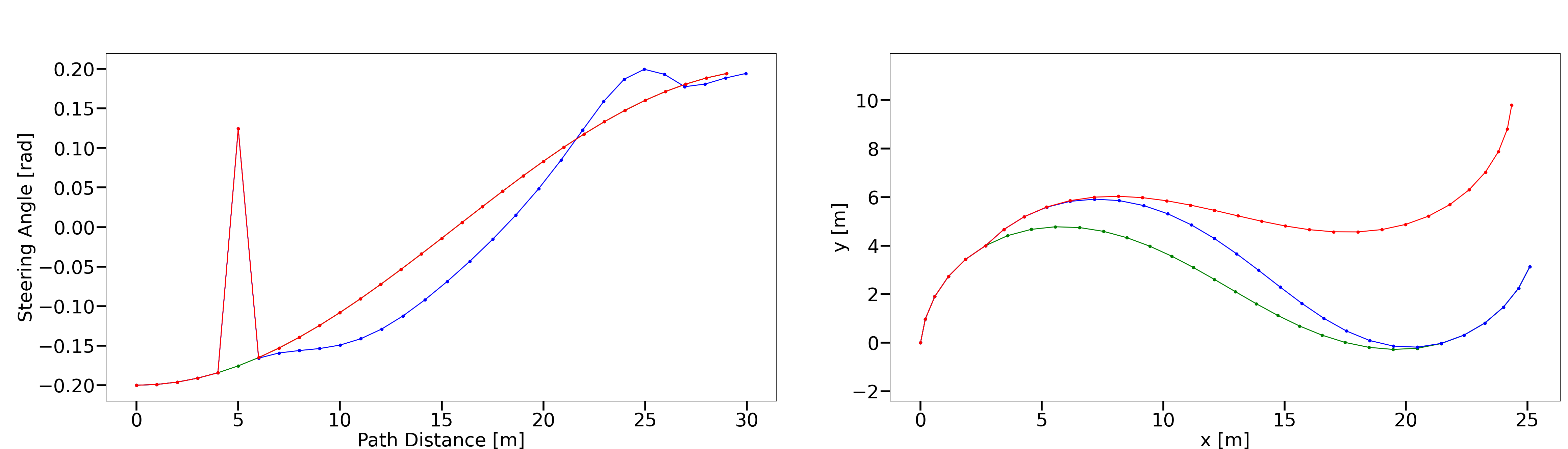}
	}}
      
      \caption{Path perturbations in actuator space (left) and world space (right). Green is the unperturbed path, red is perturbed with no correction, and blue is perturbed with correction.}
      \label{fig:path_perturbation}
   \end{figure}

With the cost function now defined the path can be sent to the solver. Here, we cause perturbations to the path at each point such that each perturbation decreases the cost of the path. However, perturbations at the start of the path will have a much larger total effect on the path than perturbations later in the path. Additionally, the actuator space does not fix the goal point of the path. To solve both these issues we dampen the perturbation over the next N time steps. This is done by using an S curve to interpolate between the perturb path and the original path in world space then transforming this path back into actuator space. Figure~\ref{fig:path_perturbation} shows this process for a perturbation at step 5 with a damping length of 25 time steps.

The solver itself is a simple implementation of gradient descent. It is described by algorithm~\ref{alg:path_solver}. In brief, the cost for the perturbed segment is checked for both left and right perturbations over the perturbation distance.  This, along with the current cost of that segment, defines the gradient. Then, a final perturbation is chosen for that point according to the gradient and a scaling factor. This process is repeated for each point along the path. When there exist no more cost reducing perturbations along the path, or the maximum iteration limit is reached, the path is returned.

\begin{algorithm}
 \caption{Optimizer}
 \begin{algorithmic}[1]
 \label{alg:path_solver}
 \renewcommand{\algorithmicrequire}{\textbf{Input:}}
 \renewcommand{\algorithmicensure}{\textbf{Output:}}
 \STATE $currentCost = Cost(path)$
 \STATE $lastCost = inf$
 \WHILE {$ currentCost < lastCost $}
  \STATE $lastCost = currentCost$
  \FOR {$point$ in $path$}
	\STATE $cost_{left} = Cost(Perturb(point,left))$
	\STATE $cost_{right} = Cost(Perturb(point,right))$
	\IF {$ cost_{left} < currentCost$}
		\STATE $ scale = currentCost - cost_{left}$
		\STATE $ path = Perturb(point,left * scale)$ 
		\STATE $ currentCost = Cost(path)$
	\ENDIF
	\IF {$ cost_{right} < currentCost$}
		\STATE $ scale = currentCost - cost_{right}$
		\STATE $ path = Perturb(point,right * scale)$ 
		\STATE $ currentCost = Cost(path)$
	\ENDIF
  \ENDFOR 
 \ENDWHILE
 \STATE return $path$ 
 \end{algorithmic} 
 \end{algorithm}

\subsection{Speed Control}
So far we have only been concerned with the steering angle of the vehicle and any obstacles the path might run into. However, we also want the vehicle to traverse this path as fast as possible. Thus the need for a speed planner that will maintain a high speed while respecting the constraints of the vehicle. Speed control is done in two passes. First, the maximum speed at each point is determined based on that point's steering angle (equation~\ref{equ:speed_controll}). Here $v_{min}$ and $v_{max}$ are the minimum and maximum speeds, $\gamma$ is the steering angle at which we want the minimum speed, and $p_{v}, p_{\phi}$ are the steering angle and speed for the point $p$. Additionally, the goal point is assigned a speed of zero.

\begin{equation}
\label{equ:speed_controll}
p_v = v_{min} + (v_{max} - v_{min}) (1 - \min(\frac{ \vert p_{\phi} \vert}{\gamma},1)) 
\end{equation}

Next, backwards and forwards passes are done over the path such that all the path speeds lie within the maximum and minimum acceleration limits of the vehicle.

After this speed control pass the path is converted back into world space and then published to the vehicle controller.

\section{RESULTS}

\subsection{System Overview}

\begin{figure}[htpb]
      \centering
      \framebox{\parbox{3in}{
      
      \includegraphics[width=\linewidth]{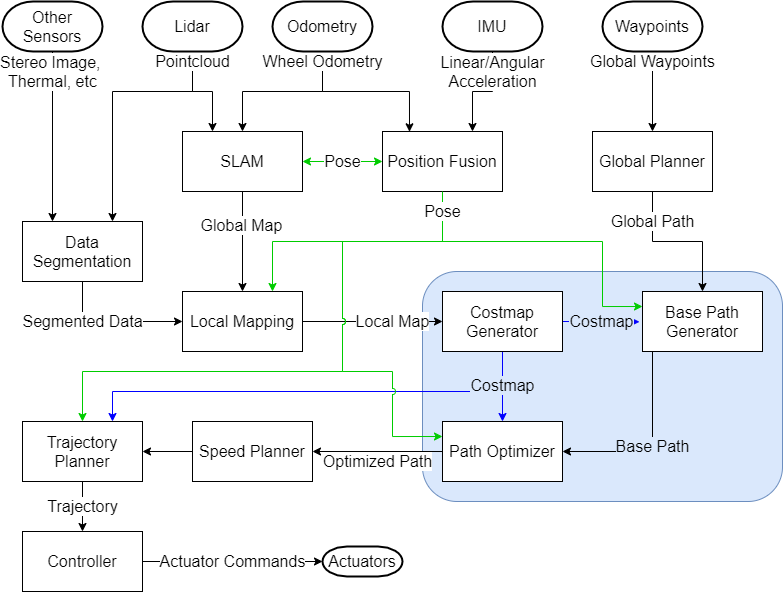}
	}}
      
      \caption{A system overview of our test setup. The shaded region represents the topic of this paper as represented in figure~\ref{fig:opt_architecture}.}
      \label{fig:system_overview}
   \end{figure}

We tested this system on a Clearpath robotics Warthog. The system architecture is described in figure~\ref{fig:system_overview}. We equipped it with an Ouster OS1-64 lidar and a Vectornav VN-300 INS. Data from the IMU and wheel odometry was fused using an extended kallman filter to provide fused odometry data. Our map had a resolution of 20~cm with a size of 512x512 and updated at a rate of 2~Hz. The A* planner was ran at 4~Hz with the optimization module running at 10~Hz. The tests were done with a speed of 5~m/s, the vehicle's maximum speed, and a minimum turning speed of 2~m/s.

\subsection{Experimental Results}

   \begin{figure}[htbp]
      \centering
      \framebox{\parbox{3in}{
      
      \includegraphics[width=\linewidth]{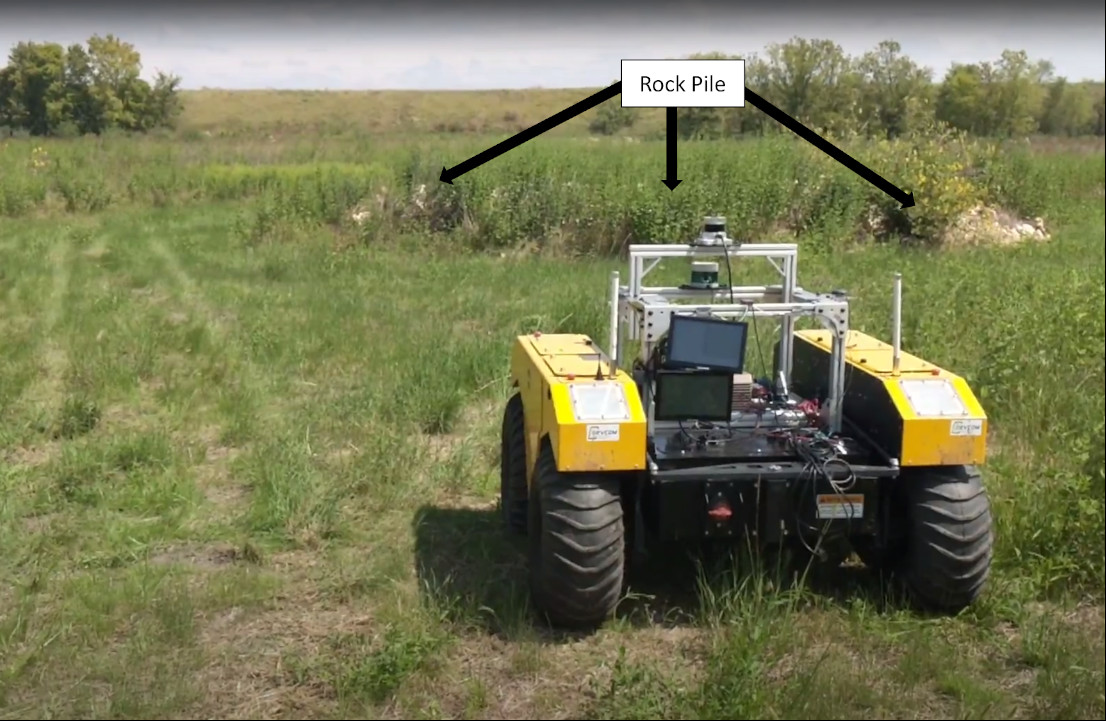}
	}}
      
      \caption{The test site used for the first experiment at the Texas A\&M RELLIS campus with a large rock pile covered in vegetation. }
      \label{fig:rock_picture}
   \end{figure}

\begin{figure}[htbp]
      \centering
      \framebox{\parbox{3in}{
      
      \includegraphics[width=\linewidth]{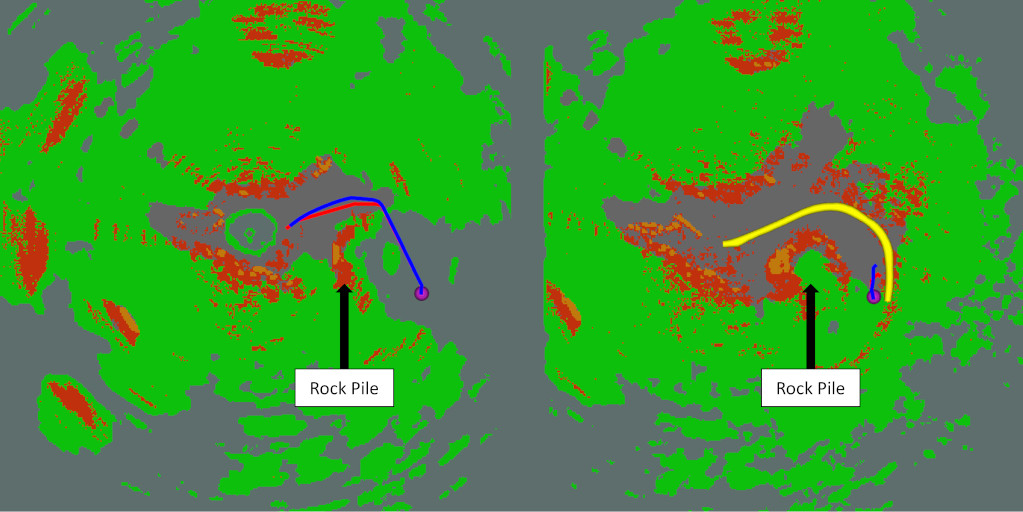}

	}}
      
      \caption{Driving around the rock pile with the initial plan (left) and final path (right). Gray is free space, green is rough ground, bright orange is hard obstacles, and dark orange is soft obstacles. The red path is the input path to the optimizer, blue is the optimized path, and yellow is the driven path. The goal point is represented by the pink circle. }
      \label{fig:rock_results}
   \end{figure}

The following experiments were done at the Texas A\&M RELLIS campus. This first test was done around a large rock pile (figure~\ref{fig:rock_picture}). Figure~\ref{fig:rock_results} shows the test on the map. The goal point was given on the other side of the pile. The line in red shows the initial A* path, the blue line shows the resulting optimized path. Note that the resulting path is similar to the input path but avoids the sharp turns in the base path. The right figure shows the final driven path in yellow. We can see that the driven path is initially similar to the first plan. However, once the vehicle can see the other side of the rocks the path is re-planned.

   \begin{figure}[h!tbp]
      \centering
      \framebox{\parbox{3in}{
      
      \includegraphics[width=\linewidth]{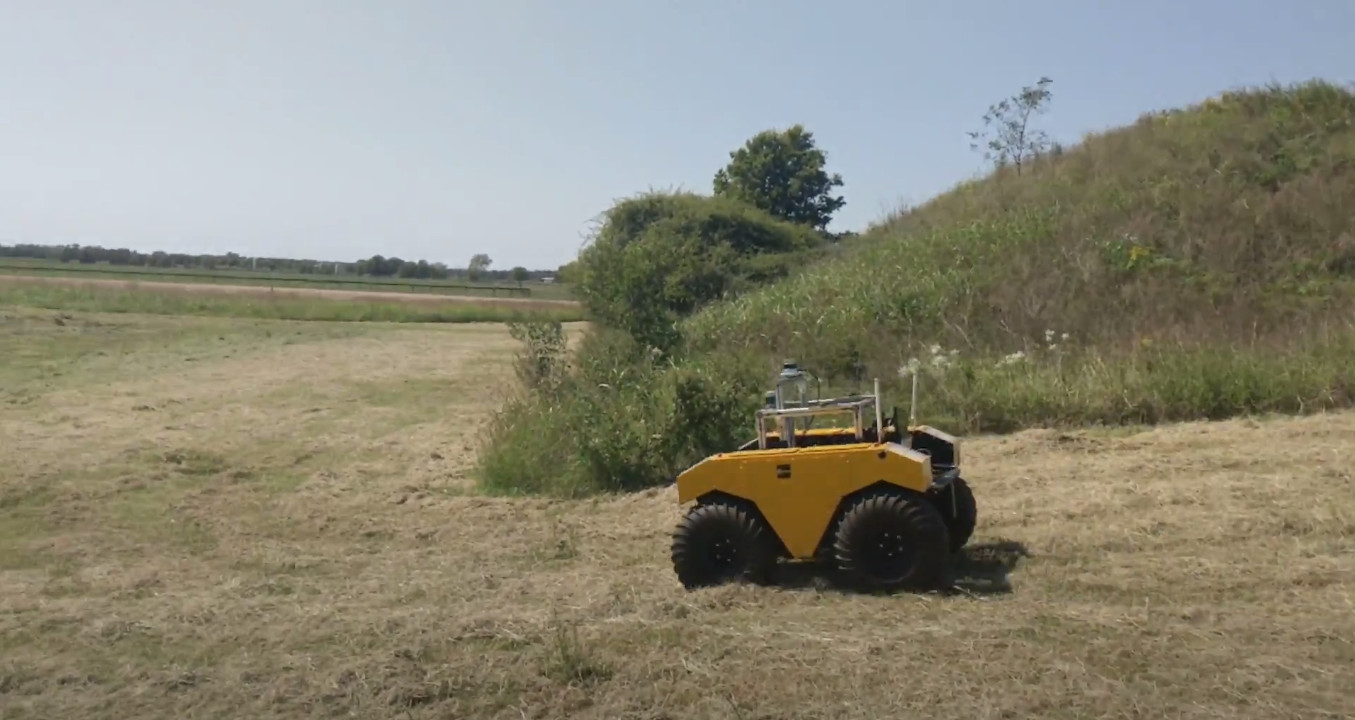}
	}}
      
      \caption{The test site used for the second experiment at the Texas A\&M RELLIS campus driving around a large hill. }
      \label{fig:hill_picture}
   \end{figure}

\begin{figure}[h!tbp]
      \centering
      \framebox{\parbox{3in}{
      
      \includegraphics[width=\linewidth]{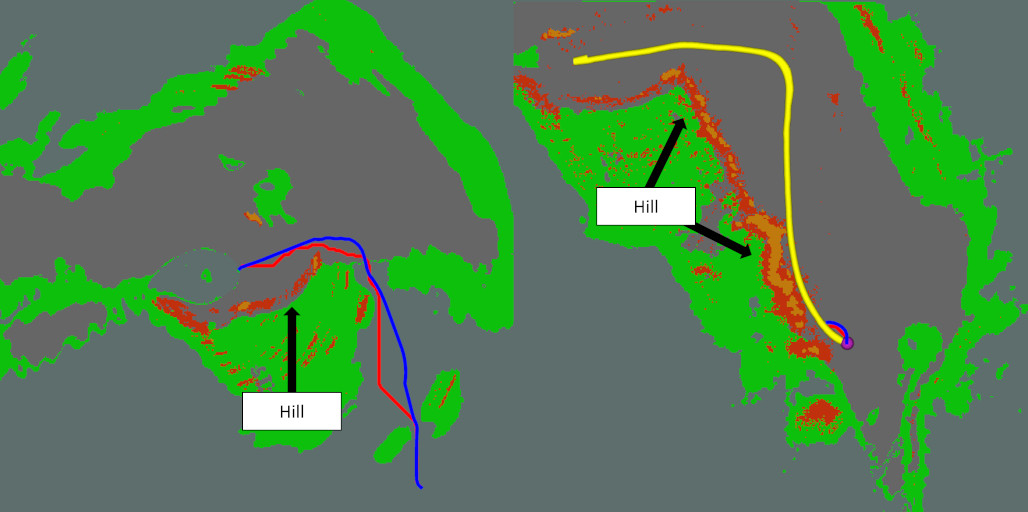}

	}}
      
      \caption{Driving around a large hill with the initial plan (left) and final path (right). The colors on the map represent different terrain features. Gray is free space, green is rough ground, bright orange is hard obstacles, and dark orange is soft obstacles. The red path is the input path to the optimizer, blue is the optimized path, and yellow is the driven path. The goal point is represented by the pink circle.}
      \label{fig:hill_results}
   \end{figure}

In test 2 we go around a large hill (figure~\ref{fig:hill_picture}). Note that the optimized path is significantly different from the base path (figure~\ref{fig:hill_results}). Unlike the first experiment, the terrain around the hill is largely empty. Here, the planned path didn't significantly change as the robot moved, but controller errors led to an overshoot of the first turn. However, this was smoothly corrected and the resulting path was still collision free.

\subsection{Simulated Results}
\begin{figure}[htbp]
      \centering
      \framebox{\parbox{3in}{
      
      \includegraphics[width=\linewidth]{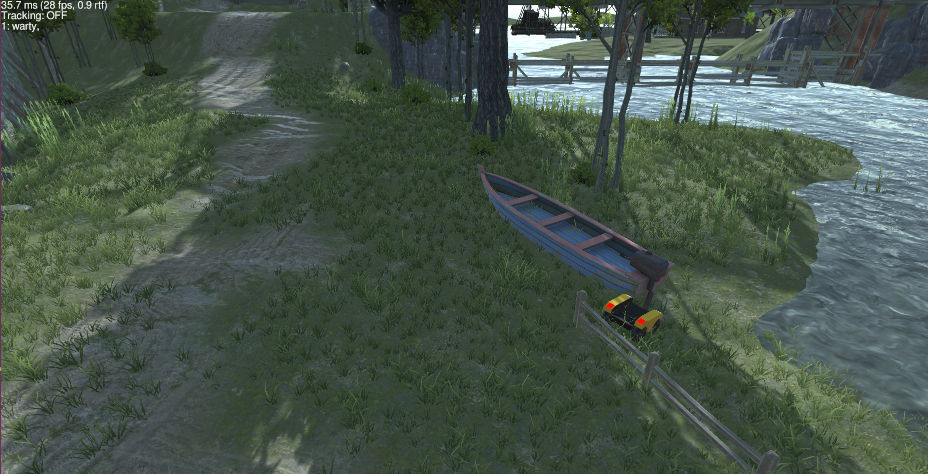}
	}}
      
      \caption{The simulation environment showing the final position of the Warthog after the A* baseline test.}
      \label{fig:sim1}
   \end{figure}

\begin{figure}[htpb]
      \centering
      \framebox{\parbox{3in}{
      
      \includegraphics[width=\linewidth]{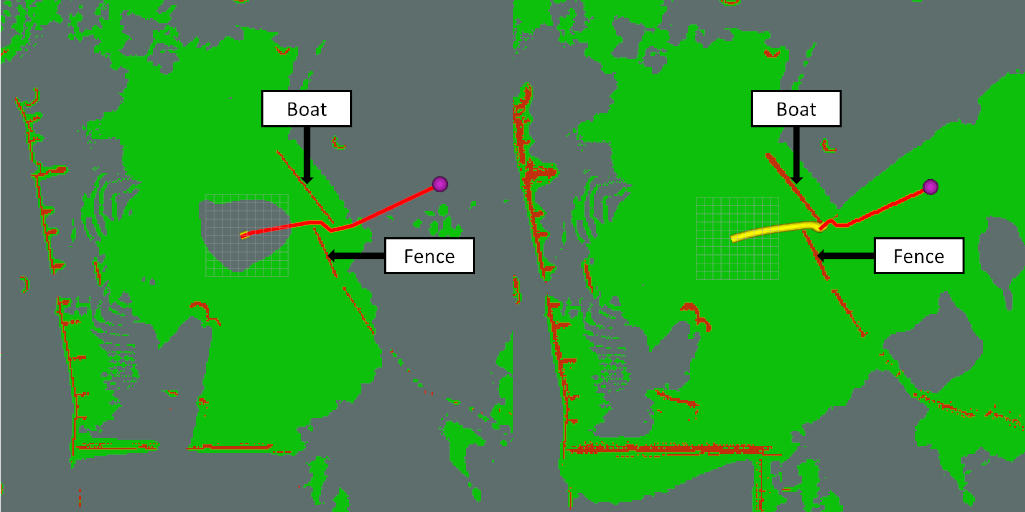}
	}}
      
      \caption{A* sim results with the initial plan (left) and driven path (right). The colors on the map represent different terrain features. Gray is unknown space, green is ground, and dark orange is obstacles. The red path is the initial a* path and yellow is the driven path. The goal point is represented by the pink circle.}
      \label{fig:sim_a_star}
   \end{figure}

\begin{figure}[h!tbp]
      \centering
      \framebox{\parbox{3in}{
      
      \includegraphics[width=\linewidth]{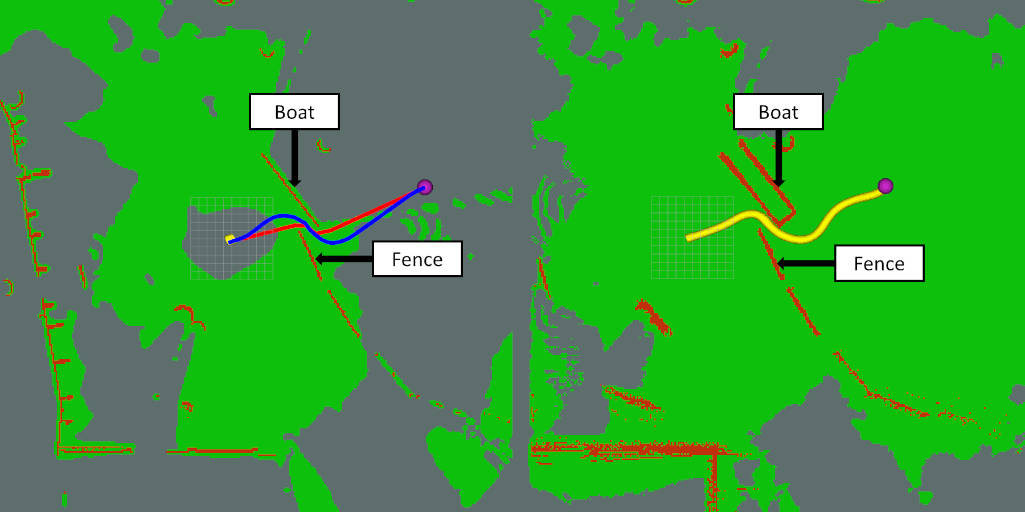}
	}}
      
      \caption{Optimized results with the initial plan (left) and driven path (right). The colors on the map represent different terrain features. Gray is unknown space, green is ground, and dark orange is obstacles. The red path is the initial a* path, blue is the optimized path, and yellow is the driven path. The goal point is represented by the pink circle.}
      \label{fig:sim_optimized}
   \end{figure}

In addition to the real testing we also did several runs in simulation. The simulator was created in the Unity game engine and is integrated with the Robot Operating System (ROS). The simulation vehicle is a replica of the Warthog used for the physical testing. Figure~\ref{fig:sim1} shows the simulated test environment. First we ran the system with the A* planner as a baseline (figure~\ref{fig:sim_a_star}). Next, the same path was ran with our optimizer running on top of the A* path (figure~\ref{fig:sim_optimized}). All other settings were left unchanged. A maximum speed of 5~m/s and a minimum speed of 2~m/s were used. The goal point was set so the vehicle would go the small gap between the fence and boat. With only the A* path the vehicle would overshoot the path and get stuck in the gap. Adding the optimizer allowed the vehicle to successfully navigate the obstacles.

\section{CONCLUSION}

We have developed a path optimization method for ground vehicles that is able to easily enforce kinematic constraints onto the path. Using a gridded A* path as a base, this planner was successfully able to operate at high speeds over off-road terrains.

Going forward we would like to extend this path optimization method to speed control and temporal cost maps. However, increasing the dimensonality of the problem would also necessitate a more robust and efficient solver. By refining the cost function better properties could be expected. These properties could then be exploited to more efficiently solve the optimization function. Although, as terrain classification gets more complicated more layers to the costmap will become necessary. This may increase the potential of local minima and will be an important consideration in future work. Additionally, planning in the actuator space presents many opportunities for a tighter coupling between planning and control. Finally, a mention should be made about planning rate with increasing speed. To maintain a constant reaction distance the total system latency must be inversely proportional to speed. Although planning is a large part of the latency, to be successful at high speed operation the entire system must be considered.

\balance



\bibliographystyle{IEEEtran}
\bibliography{IEEEabrv,IEEEexample}

\end{document}